\documentclass{article} % For LaTeX2e
\usepackage{iclr2026_conference,times}

\usepackage[utf8]{inputenc}
\usepackage[english]{babel}
\usepackage{amsmath}
\usepackage{amsfonts}
\usepackage{amssymb}
\usepackage{graphicx}
\usepackage{hyperref}
\usepackage{geometry}
\usepackage{booktabs}
\usepackage{caption}
\usepackage{xcolor}
\usepackage{tabularx}
\usepackage{}

\hypersetup{
    colorlinks=true,
    linkcolor=blue,
    filecolor=magenta,      
    urlcolor=cyan,
    pdftitle={LinguistAgent Technical Report},
}

\title{\textbf{LinguistAgent Technical Report: \\ A Reflective Multi-Model Platform for Automated Linguistic Annotation}}
\author{
\makebox[\textwidth]{\textbf{Bingru Li}} \\
\makebox[\textwidth]{\small University of Birmingham} \\
\makebox[\textwidth]{\small \texttt{bxl426@student.bham.ac.uk}}
}

\date{January 2025}
\iclrfinalcopy % Uncomment for camera-ready version, but NOT for submission.

\begin{document}

\maketitle

\begin{abstract}
Data annotation remains a significant bottleneck in the field of humanities and social sciences, particularly for complex linguistic tasks such as metaphor identification. While Large Language Models (LLMs) show promise, a significant gap remains between the theoretical capability of LLMs and their practical utility for researchers. This paper introduces \textbf{LinguistAgent}, an integrated, user-friendly platform that leverages a reflective multi-model architecture to automate linguistic annotation. The platform comprises an \textit{Annotator} and an optional \textit{Reviewer} to simulate a peer-review process. This platform supports comparative experiments across three main paradigms: Prompt Engineering (Zero-shot/Few-shot/Chain-of-thought), Retrieval-Augmented Generation, and Fine-tuning. We demonstrate LinguistAgent's efficacy by replicating the task of metaphor identification from a published study, which provides real-time token-level evaluation ($F_1$ and Cohen's kappa) against human gold standards. The application and codes are released on \url{https://github.com/Bingru-Li/LinguistAgent}.
\end{abstract}

\section{Introduction}
Linguistic annotation often requires deep contextual and logical reasoning. A long-standing case is the identification of metaphors in discourse \cite{johnson1980metaphors, cameron2003metaphor, cameron2010responding,steen2009three, cameron2010metaphor, littlemore2020metaphors, fuoli2022sunken, turner2023literal}, which demands intensive human labor. Both the Metaphor Identification Procedure (MIP) \cite{group2007mip,steen2010method} and the Procedure for Identifying Metaphorical Scenes (PIMS) \cite{johansson2023procedure} involve comparing the contextual meaning of a word with its basic, more concrete meaning. Manually applying these protocols to large-scale corpora is time-consuming. 

Recent benchmarks \cite{ge2023survey,tian2024theory} suggest that Large Language Models (LLMs) can achieve near-human reliability. However, a significant gap remains between the theoretical capability of LLMs and their practical utility. Current computational approaches typically rely on prompt-based approaches \cite{puraivan2024metaphor,liang2025using,hicke2024science}, which often oversimplifies the multi-stage reasoning inherent in protocols like MIP. Conversely, more sophisticated pipelines \cite{fuoli2026metaphor} often require extensive programming knowledge, making them less accessible to non-expert researchers. {\textbf{LinguistAgent}} bridges this gap by providing a no-code, multi-agent environment designed specifically for large-scale data annotation and benchmarking.

\begin{figure}[h]
    \centering
    \includegraphics[width=\linewidth]{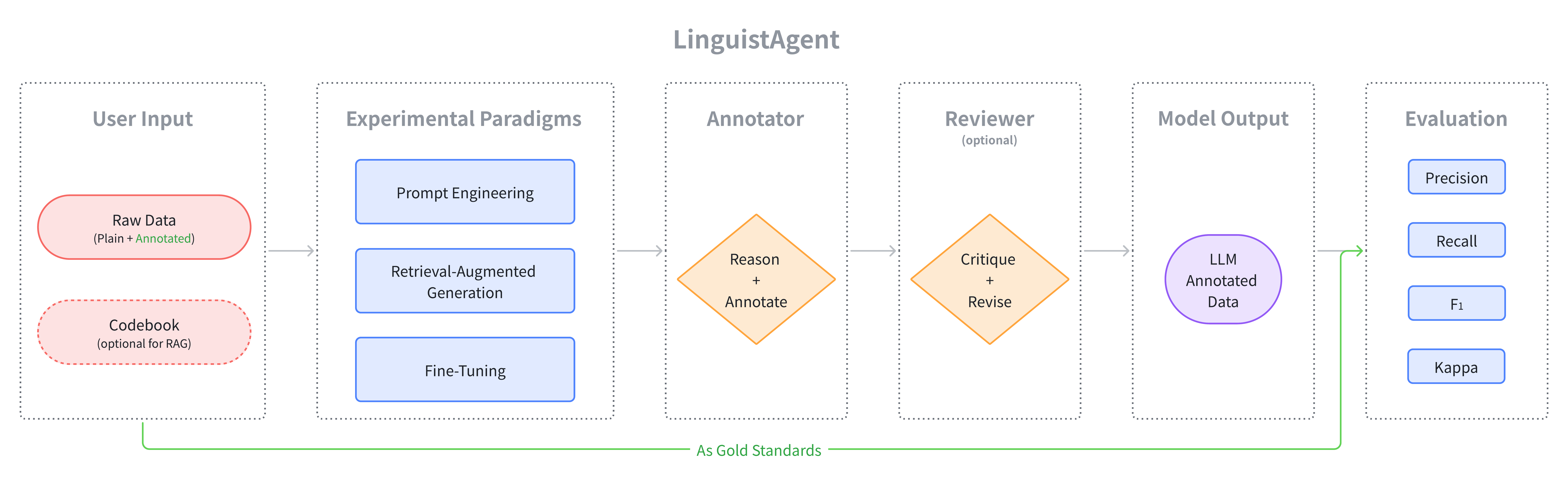}
    \caption{The workflow of LinguistAgent.}
    \label{fig:flowchart}
\end{figure}

\begin{figure}[h]
    \centering
    \includegraphics[width=\linewidth]{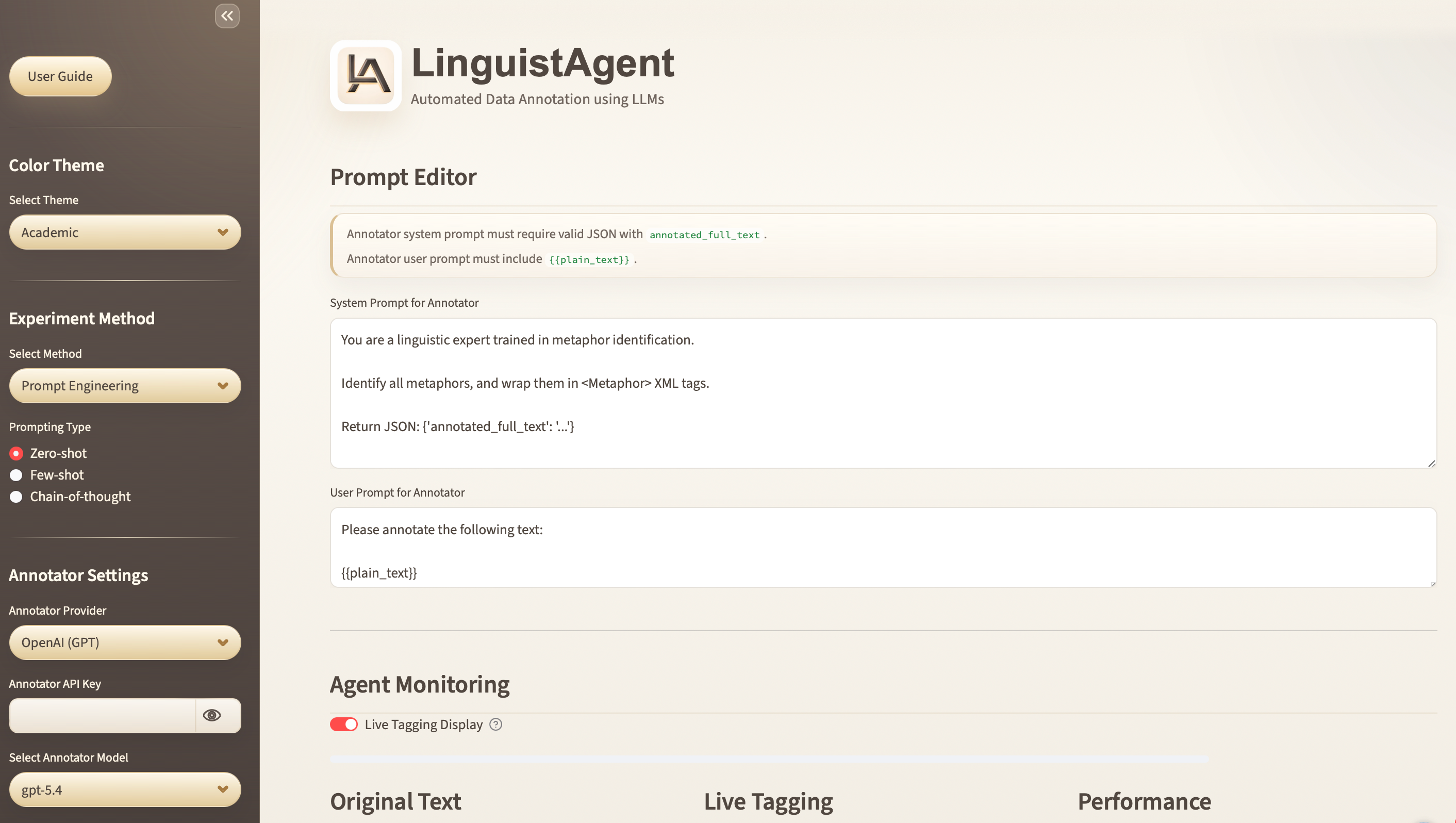}
    \caption{The user interface of LinguistAgent.}
    \label{fig:ui-raw}
\end{figure}

Fig. \ref{fig:flowchart} illustrates the workflow of the platform, where the user is expected to upload the raw data, including plain text and optional human annotated text as gold standards. Under a specific experimental paradigm, i.e., Prompt Engineering, Retrieval-Augmented Generation (RAG), and Fine-tuning, the Annotator LLM will code the data as instructed. If the Reviewer Mode is turned on, another Reviewer LLM will check and revise the first round of annotations when necessary. The final LLM annotated data will be compared against the uploaded gold standards on such evaluation metrics as precision, recall, $F_1$, and Cohen's kappa ($k$). The default user interface (UI) is shown in Fig. \ref{fig:ui-raw}.

\section{The Functions of LinguistAgent}

\subsection{The Annotator Agent}
The Annotator acts as the primary executor. It receives the raw text and an optional \textit{Codebook} (in the RAG scenario), and its task is to perform sequence labeling by wrapping target expressions in XML tags (e.g., \texttt{<Metaphor>...</Metaphor>}). The platform supports various model providers, including leading closed-source providers such as OpenAI (GPT), Google (Gemini), and open-source providers such as DeepSeek, Alibaba (Qwen).

\subsection{The Reviewer Agent}
The Reviewer can be activated by turning on Reviewer Mode and choosing another model, which serves to check the annotations. It receives the original raw data and the Annotator's annotated data, identifying false positives or missed instances. If discrepancies are found, the Reviewer provides a \textit{Critique} and generates a \textit{Revised Text}.

% \begin{figure}[h]
%     \centering
%     \includegraphics[width=\linewidth]{figure/reviewer.jpg}
%     \caption{Reviewer critique.}
%     \label{fig:reviewer}
% \end{figure}

\subsection{Experimental Methods}
\begin{figure}[h]
    \centering
    \includegraphics[width=\linewidth]{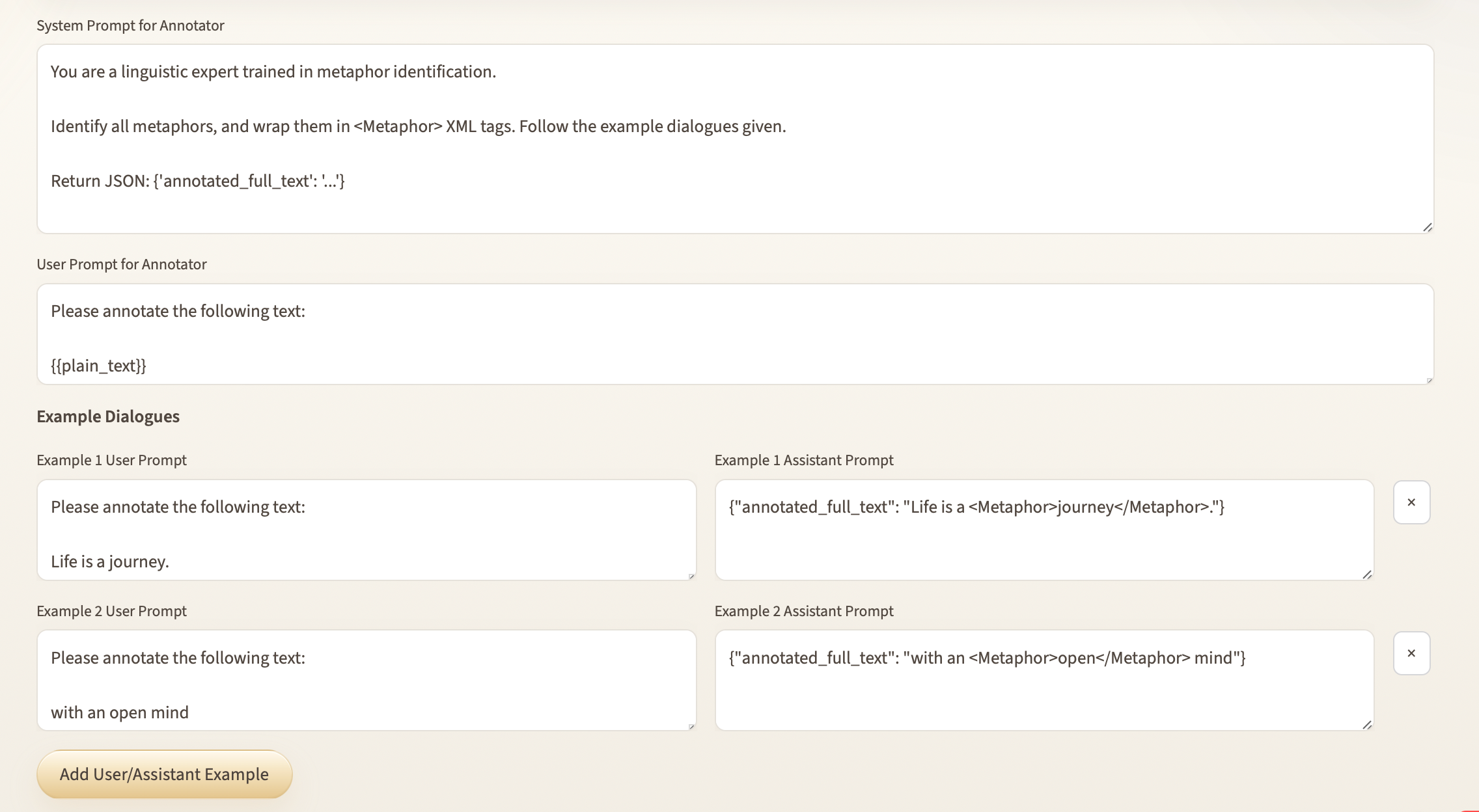}
    \caption{Few-shot/Chain-of-thought prompting allows one to add User/Assistant prompt examples.}
    \label{fig:few-shot}
\end{figure}

The platform integrates three distinct LLM application strategies to allow researchers to find the optimal balance between cost and accuracy:
\begin{itemize}
    \item \textbf{Prompt Engineering} supports \textit{Zero-shot} prompting, which only contains task instructions, and \textit{Few-shot} and \textit{Chain-of-thought} prompting, where User/Assistant prompt examples can be added for the LLMs to learn from, as shown in Figure \ref{fig:few-shot};
    \item \textbf{RAG} enables excerpt retrieval from an uploaded codebook using an embedding model, as a reference for the Annotator;
    \item \textbf{Fine-tuning} allows users to either input a \textit{Tuned Model ID}, or create a new fine-tuning job. This enables the evaluation of specialized models trained on domain-specific data, often resulting in higher performance.
\end{itemize}

\begin{figure}
    \centering
    \includegraphics[width=\linewidth]{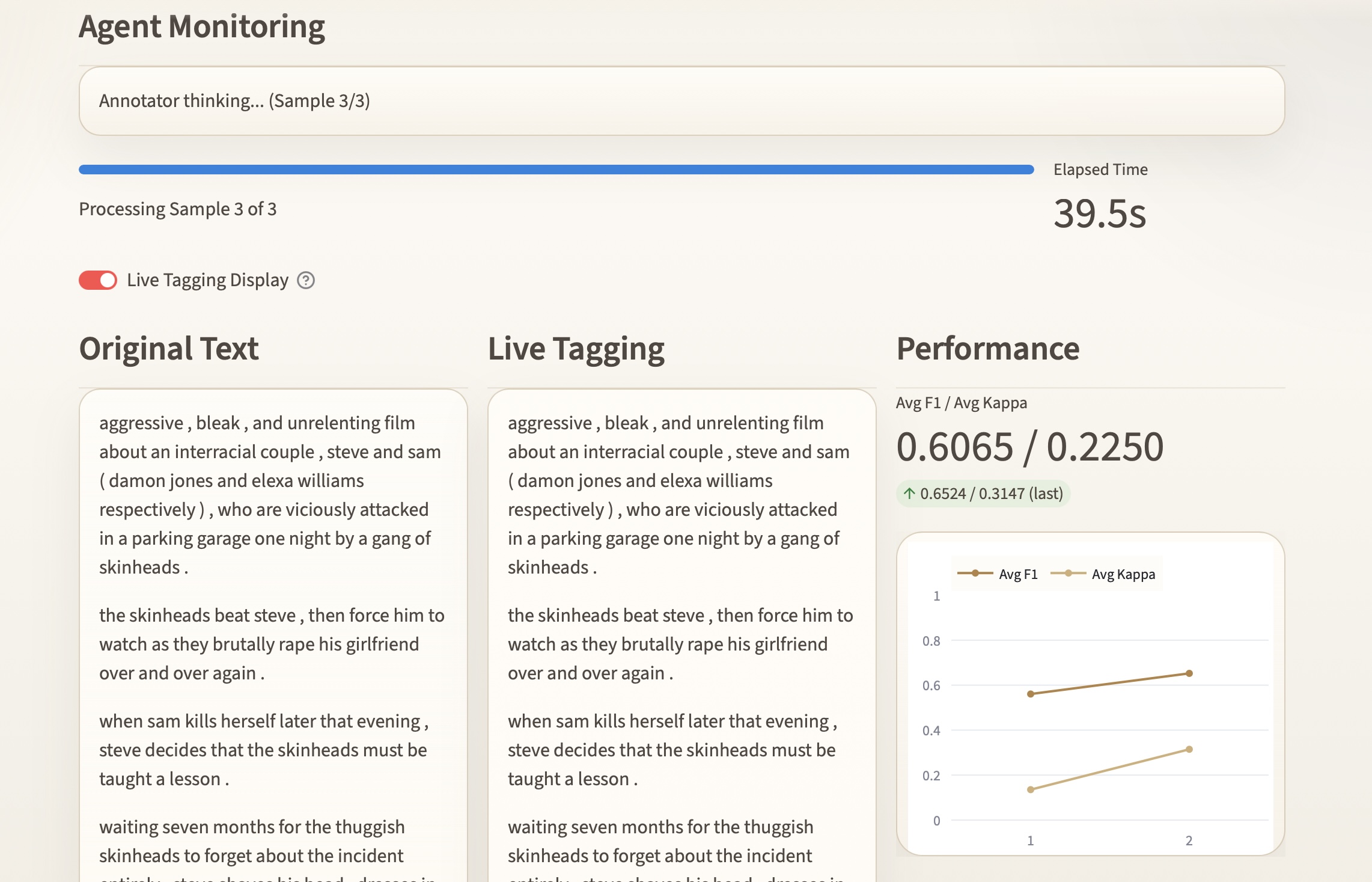}
    \caption{The Agent Monitoring section.}
    \label{fig:live}
\end{figure}

\subsection{Agent Monitoring}
Live performance and a progress bar are displayed during the experiment, as shown in Fig. \ref{fig:live}. Live tagging display can be turned off for faster experiments on large-scale datasets.  LinguistAgent also performs real-time \textbf{Token-level Evaluation}. It tokenizes both the human gold standards and the LLM predictions into binary sequences where $1$ represents a tagged unit and $0$ represents a non-tagged unit.

The following metrics are calculated for each sample: 
\begin{itemize}
    \item \textbf{Precision ($P$):} The ratio of correctly predicted metaphorical tokens to the total predicted metaphorical tokens.
    \begin{equation}
    P = \frac{TP}{TP + FP}
    \end{equation}
    \item \textbf{Recall ($R$):} The ratio of correctly predicted metaphorical tokens to all metaphorical tokens in the gold standard.
    \begin{equation}
    R = \frac{TP}{TP + FN}
    \end{equation}
    \item \textbf{F1 Score ($F_1$):} The harmonic mean of Precision and Recall.
    \begin{equation}
    F_1 = 2 \cdot \frac{P \cdot R}{P + R}
    \end{equation}
    \item \textbf{Cohen's Kappa ($\kappa$):} A chance-corrected measure of
    agreement between the model's token-level labels and the gold standards,
    where each token is labelled as metaphorical or non-metaphorical. It is
    defined as
    \begin{equation}
    \kappa = \frac{p_o - p_e}{1 - p_e}
    \end{equation}
    where $p_o$ is the observed agreement and $p_e$ is the agreement expected
    by chance. Let $N = TP + FP + FN + TN$ be the total number of tokens. The
    observed agreement is the proportion of tokens on which the two labellings
    coincide,
    \begin{equation}
    p_o = \frac{TP + TN}{N},
    \end{equation}
    and the expected agreement is computed from the marginal proportions of
    metaphorical and non-metaphorical labels assigned by each side,
    \begin{equation}
    p_e = \frac{(TP + FP)(TP + FN) + (FN + TN)(FP + TN)}{N^2}.
    \end{equation}
\end{itemize}

\subsection{Traceability and Error Analysis}
\begin{figure}[h]
    \centering
    \includegraphics[width=\linewidth]{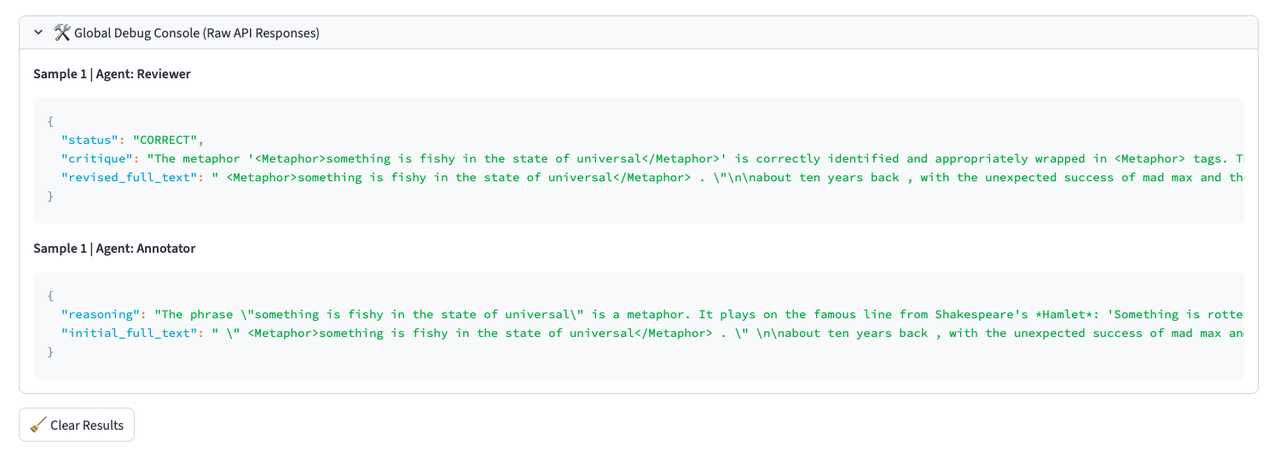}
    \caption{The debug section, keeping logs of all raw model responses and error types.}
    \label{fig:debug}
\end{figure}

A key challenge in LLM annotation is the potential for silent failures. LinguistAgent mitigates this through an integrated monitoring and debugging function:

\begin{itemize}
    \item \textbf{Real-time Reasoning Logs:} Both the Annotator and the Reviewer allows the user to require extra reasoning outputs, i.e., the reasoning of the Annotator and the critique of the Reviewer. This provids a window into the agents' linguistic logic before the final labels are applied.
    \item \textbf{Failure Diagnosis:} The Debug Console enables the identification of specific structural issues (see Figure \ref{fig:debug}), such as the model reaching the \texttt{max\_output\_tokens} limit (resulting in a truncated JSON) or the API quota limit. 
    \item \textbf{Full Records of Everything:} LinguistAgent generates a downloadable CSV containing the original data, model information, model annotations, and per-sample evaluation metrics (before and after review), to assist in further in-depth manual investigation. 
\end{itemize}
This level of transparency ensures verifiable outcomes of a transparent workflow, allowing for in-depth analysis of the LLM annotation.

\section{Implementation and System Observability}

The current version of LinguistAgent is developed as a web-based application designed to bridge the gap between complex AI workflows and intuitive research interfaces. By providing a user-friendly, code-free environment, LinguistAgent empowers non-technical scholars to harness the potential of LLMs for their specific research needs without profound programming knowledge. This accessibility facilitates a more inclusive research ecosystem.

\subsection{Frontend Architecture: Streamlit Integration}
The application frontend is built using the \textbf{Streamlit} framework, chosen for its ability to create highly reactive user interfaces. 
\begin{itemize}
    \item \textbf{Dynamic Configuration Interface:} The UI provides a configuration suite where researchers can dynamically switch between experimental methods (Prompt Engineering, RAG, and Fine-tuning). It supports independent model and API key selection for each agent role, allowing for heterogeneous workflows.
    \item \textbf{Reactive State Management:} We leveraged the \textit{Session State} of Streamlit to ensure persistence of experimental data. This allows the dashboard, including the live tagging view and the performance charts, to remain interactive even after a batch process triggers a page rerun to enable data export.
    \item \textbf{Real-time Visualization:} LinguistAgent integrates \textbf{Plotly} for dynamic performance tracking. As each sample is processed, token-level $F_1$ and $k$ scores are plotted in real-time against an adjustable academic baseline, providing immediate visual feedback on the model's reliability.
\end{itemize}

\subsection{Backend Logic and Model Heterogeneity}
The backend is engineered to handle various model architectures through a unified communication wrapper.
\begin{itemize}
    \item \textbf{Multi-Provider Support:} LinguistAgent standardizes interactions with the Google GenAI SDK and OpenAI-compatible endpoints. This allows for the simultaneous use of disparate models, such as using Alibaba's Qwen for initial annotation and Google's Gemini for the high-reasoning Reviewer role.
    \item \textbf{Structured Output Control:} We enforce Native JSON Mode across all providers to ensure that reasoning chains and annotations are programmatically separated, which is vital for the automated evaluation engine.
    % \item \textbf{Permissive Safety Configuration:} For scientific research involving film reviews or social media discourse, safety thresholds are set to \texttt{BLOCK\_NONE} to ensure that valid linguistic data is not filtered out by the model's default safety policies.
\end{itemize}

\section{Case Study: Metaphor Identification}
We conducted a pilot replication study on the metaphor identification task \cite{fuoli2026metaphor} using GPT-4.1 on LinguistAgent, and compared the Annotator-only scenario and the Reviewer Mode. Preliminary results are shown in Table \ref{tab: metaphor_f1}. Although we tried to keep all experimental conditions the same for the purpose of replication, an inevitable difference is that the platform requires a JSON format output from the LLMs to enable real-time evaluation. For RAG, however, we used three embeddings models (text-embedding-3-small, text-embedding-3-large, and TF-IDF) that are different from the one used in the original paper, which is a local embedding mode (nomic-embed-text) that requires extra programming from the user. For fine-tuning, we incorporated a simple system prompt at the inference stage after the fine-tuning job, which may explain the rise of $F_1$ in this scenario. These pilot results indicate that the platform can reliably replicate the task with acceptable variances.

As for the Reviewer Mode, it is surprising that, simple as the Reviewer prompt is, it can improve overall performance in all scenarios besides Fine-tuning, be it based on self-review (GPT-4.1 vs. GPT-4.1), or independent review (GPT-4.1 vs. o1). 

\begin{table}[h]
  \caption{Metaphor identification using GPT-4.1, based on Average F1.}
  \label{tab: metaphor_f1}
  \centering
  \small
  \begin{tabularx}{\textwidth}{Xccccc}
    \toprule
    \textbf{Model $\times$ Paradigm} & Zero-shot & Few-shot & CoT & RAG & FT\\
    \midrule
    \multicolumn{4}{l}{\textit{Original Results}} \\ % 这里是第一组的副标题
    \midrule
    GPT-4.1 & 0.596   & 0.603   & 0.658   & 0.608  & 0.765  \\
    \midrule
    \multicolumn{4}{l}{\textit{LinguistAgent: Annotator Only}} \\ % 这里是第二组的副标题
    \midrule
    GPT-4.1 & 0.589  & 0.614  & 0.635  & 0.637 & 0.822 \\
    \midrule
    \multicolumn{4}{l}{\textit{LinguistAgent: Reviewer Mode On}} \\ % 这里是第三组的副标题
    \midrule
    GPT-4.1 & 0.656  & 0.679  & 0.682 & 0.671 & 0.791  \\ % 示例数据
    o1 (reasoning) & 0.666  & 0.661  & 0.670 & 0.662  & 0.765   \\ % 示例数据
    \bottomrule
  \end{tabularx}
\end{table}

\section{Conclusion}
LinguistAgent demonstrates that agentic workflows can transform LLMs into rigorous scientific instruments. By automating the ``Annotate-Review-Evaluate'' cycle, the platform empowers researchers in the Humanities and Social Sciences to scale their analyses while maintaining high levels of transparency and replicability. Future work should focus on integrating ``Human-in-the-loop'' verification to further refine the platform.

\bibliographystyle{iclr2026_conference}
\bibliography{references}

% \begin{thebibliography}{9}
% \bibitem{fuoli2025}
% Fuoli, M., Huang, W., Littlemore, J., Turner, S., \& Wilding, E. (2025). \textit{Metaphor Identification Using Large Language Models: A Comparison of RAG, Prompt Engineering, and Fine-Tuning.} arXiv preprint.

% \bibitem{gemini2024}
% Google Gemini Team. (2024). \textit{Gemini 1.5: Unlocking multimodal understanding across millions of tokens of context.} Technical Report.

% \bibitem{steen2010}
% Steen, G. J., et al. (2010). \textit{A Method for Linguistic Metaphor Identification: From MIP to MIPVU.} John Benjamins Publishing.
% \end{thebibliography}

\end{document}